\documentclass{article}
\usepackage[nonatbib, preprint]{neurips_2023}
\usepackage{lipsum}
\usepackage{dirtytalk}
\usepackage{todonotes}
\usepackage{array,multirow}
\usepackage{booktabs}       % professional-quality tables
\usepackage{amsfonts}       % blackboard math symbols
\usepackage{nicefrac}       % compact symbols for 1/2, etc.
\usepackage{microtype}      % microtypography
\usepackage{amsmath}
\usepackage{graphicx}
\usepackage{amssymb}
\usepackage[utf8]{inputenc} % allow utf-8 input
\usepackage[T1]{fontenc}    % use 8-bit T1 fonts
\usepackage{hyperref}       % hyperlinks
\usepackage{url}            % simple URL typesetting
\usepackage{booktabs}       % professional-quality tables
\usepackage{amsfonts}       % blackboard math symbols
\usepackage{lipsum}
\usepackage{graphicx}
\graphicspath{ {./images/} }
\usepackage{array,multirow}
\usepackage{amsmath}
\usepackage{amssymb}
\usepackage{mathtools}
\usepackage{dirtytalk}

\usepackage{pifont}
\usepackage{subcaption}
\title{Privacy-Utility Trade-offs in Neural Networks for Medical Population Graphs: Insights from Differential Privacy and Graph Structure}
\author{Tamara T. Mueller*\footnotemark[3] , Maulik Chevli*\footnotemark[3] , Ameya Daigavane\footnotemark[2] , Daniel Rueckert\footnotemark[3] , Georgios Kaissis\footnotemark[3] \\ \\
Technical University of Munich\footnotemark[3] , \; Massachusetts Institute of Technology\footnotemark[2] , \; equal contribution*}

\begin{document}

\maketitle

\begin{abstract}
    %Differential privacy (DP) is the gold standard for protecting individuals' data while enabling deep learning. It is well established and frequently used for applications in medicine and healthcare to protect sensitive patient data. When using graph deep learning on so-called population graphs however, the application of DP becomes more challenging compared to grid-like data structures like images or tables.
    %The application of DP to DL methods on non-Euclidean datasets with graph deep learning holds significant additional challenges compared to grid-like data structures like images or tabular data. 
    We initiate an empirical investigation into differentially private graph neural networks on population graphs from the medical domain by examining privacy-utility trade-offs at different privacy levels on both real-world and synthetic datasets and performing auditing through membership inference attacks. 
    Our findings highlight the potential and the challenges of this specific DP application area. 
    Moreover, we find evidence that the underlying graph structure constitutes a potential factor for larger performance gaps by showing a correlation between the degree of graph homophily and the accuracy of the trained model.
\end{abstract}

\section{Introduction}
% \subsection{GNNs and Population Graphs}
Graph neural networks (GNNs) are powerful methods to apply deep learning (DL) to non-Euclidean datasets, like graphs or meshes \cite{bronstein2017geometric}.
A graph $\mathit{G}:=(\mathit{V}, \mathit{E})$ is defined as a set of vertices/nodes $\mathit{V}$ and a set of edges $\mathit{E}$, connecting nodes. 
A neighbourhood of a node $v \in \mathit{V}$ contains all nodes $u \in \mathit{V}$ for which an edge $e_{uv}$ from $u$ to $v$ exists.
GNNs follow a \textit{message passing} scheme, where node features are shared and aggregated across neighbourhoods of $n$ hops \cite{kipf2016semi,chiang2019cluster,huang2020combining,kong2020flag}.
An $n$-layer GNN propagates the information that is stored in the node features across $n$-hop neighbourhoods.
The utilisation of GNNs has shown improved performance, even on dataset not exhibiting an intrinsic graph structure, e.g. 3D point clouds \cite{wang2019dynamic}.
The application of GNNs has therefore been expanded to datasets which require constructing a graph structure prior to learning.
One such example are population graphs \cite{parisot2017spectral}.
Here, a cohort is represented by one (typically large) graph, in which nodes represent subjects and edges connect similar subjects.
The construction of the graph's structure is an important step in this pipeline, since a graph of poor structural quality can subsantially hinder graph learning \cite{luan2022we,zhu2020beyond}.
This has been attributed to several different graph properties, one of them being \textit{homophily}. 
Homophily is a measure for the ratio between same-labelled and differently labelled neighbours in a graph. 
A high homophily indicates that the majority of nodes in all neighbourhoods in the graph share the same label as the node of interest. 
% These population graphs hold highly sensitive medical information about every subject, which requires protection during DL. 
% \subsection{DP GNNs}

\paragraph{Motivation and Prior Work}
The application of differential privacy (DP) to graph neural networks for node classification tasks presents two main challenges:
(1) The connection and information exchange between data points requires specific definitions of DP for graph-structured data;
(2) In addition, the unboundedness of neighbourhoods in a graph makes privacy amplification by sub-sampling non-trivial.

In tabular datasets, individual data points can be treated separately.
This is not the case in graph learning settings, where nodes are connected to each other and share information \cite{mueller2023differential}.
%Since individual data points are interconnected in a graph setting, they cannot be treated individually like in tabular datasets \cite{mueller2023differential}, which
This contradicts the principle of \textit{per-sample gradients} in DP methods and thus requires specialised notions of DP on graphs \cite{mueller2023differential}.
In this work, we focus on \textit{node-level DP}, which protects the sensitive information stored in the node features of population graphs as well as the connections between neighbouring nodes. %Alternative DP definitions on graphs can guarantee \textit{edge-level DP}, protecting the information that lies in the graph's edges, or \textit{graph-level DP} protecting the graph as a whole. 
So far, only few works have investigated DP training of GNNs.
Daigavane et al. \cite{daigavane2021node} were among the first to introduce a privacy amplification by sub-sampling technique for multi-layer GNN training with DP-stochastic gradient descent (DP-SGD).
In order to enable a sensitivity analysis for DP-SGD in multi-layer GNNs, the authors apply a graph neighbourhood sampling scheme.
Here, the number of $k$-hop neighbours is bounded to a maximum node degree.
This ensures that the learned feature embeddings over the course of training are influenced by at most a bounded number of nodes.
Furthermore, the standard privacy amplification by sub-sampling technique for DP-SGD is extended, such that a gradient can depend on multiple subjects in the dataset:
First, a local $k$-hop neighbourhood of each node with a bounded number of neighbours is sampled.
Next, a subset $\mathcal{B}_t$ of $n$ sub-graphs is chosen uniformly at random from the set of sub-graphs that constitute the training set.
On these sub-samples (\say{mini-batches}), standard DP-SGD is applied by clipping the gradients, adding noise and using the noisy gradients for the update steps.
The noise is hereby calibrated to the sensitivity with respect to any individual node, which has been bounded via sub-sampling the input graph. 
The authors of \cite{daigavane2021node} show generally good performance at various privacy levels, motivating us to adopt their technique for this work.
A different line of work by Sajadmanesh et al. \cite{sajadmanesh2022gap} also introduces a method for DP training of multi-layer GNNs via \textit{aggregation perturbation}, i.e. by adding noise to the aggregation function of the GNN. 
This method can be used to ensure either node-level or edge-level DP and also ensures privacy guarantees at inference.
We intend to explore this technique as part of ongoing work.

DP with sufficiently strong guarantees naturally protects against membership inference attacks (MIAs), which aim to infer whether a certain individual was part of the training set or not.
There are a few works investigating MIAs on GNNs \cite{wu2021adapting,he2021node,duddu2020quantifying,olatunji2021membership}.
These and works on other privacy attacks on GNNs such as \textit{link stealing} attacks \cite{wu2022linkteller,he2021stealing} and \textit{inference attacks} \cite{zhang2022inference,zhang2021graphmi} highlight an increased vulnerability of GNNs compared to non-graph machine learning applications.
In this work, we extend the state-of-the art MIA technique by \cite{Carlini2021MembershipIA} to GNNs for the purpose of empirically validating the privacy guarantees.

\paragraph{Contributions}
In this work, we investigate the privacy-utility trade-offs of DP GNN training on medical population graphs.
Our contributions are as follows:
(1) To the best of our knowledge, our work demonstrates the first successful application of DP-SGD to multi-hop GNNs in medical population graphs;
(2) we empirically investigate the success of membership inference attacks (MIAs) at different levels of privacy protection and
(3) analyse the interplay between graph structure and the accuracy of DP GNNs, highlighting homophily as a key factor influencing model utility. 

\section{Experiments and Results}
\label{sec:experiments}

% \subsection{Datasets}
All experiments are performed based on the node-level DP GNN implementation of \cite{daigavane2021node}, using graph convolutional networks (GCNs) \cite{kipf2016semi} and the transductive learning approach.
We recall that transductive learning means that all node features and edges are included in the forward pass, but only the training labels are used for backpropagation.
As a baseline, we also train a multi-layer perceptron (MLP) for specific experiments below.
We use three medical datasets which are frequently used in the context of population graphs \cite{parisot2017spectral}:
The TADPOLE dataset studies Alzheimer's disease and functions as a benchmark dataset for population graphs \cite{parisot2017spectral,cosmo2020latent}.
We also use an in-house COVID dataset as a realistic, noisy medical dataset, with the task of predicting whether a COVID patient will require intensive care unit (ICU) treatment and the ABIDE dataset from the autism brain imaging data exchange \cite{di2014autism}, where we perform a binary classification task.
The ABIDE dataset is highly challenging, and therefore lends itself to investigating the impact of the graph structure on our experiments.
We therefore report results on the ABIDE dataset for two different graph structures, constructed using either $5$ or $30$ neighbours.
Furthermore, we evaluate our experiments on a synthetically generated binary classification dataset to investigate the impact of different graph structures on the performance of DP population graphs under controlled conditions.
All graph structures are generated using the $k$-nearest neighbour approach \cite{lu2022nagnn}.
Here, $k$ is a hyperparameter, which specifies how many neighbours each node has, and the $k$ most similar nodes are connected.
Details about the datasets as well as all $\delta$ values used for DP-SGD training are summarised in Table \ref{tab:homophily_delta}.

\begin{table}[h]
    \centering
    \scriptsize
    \caption{Summary of the homophily of the utilised graph structures as well as the $\delta$ values for all datasets.}
    \begin{tabular}{llll}
    \toprule
         \textbf{Dataset} & \textbf{\# Nodes} & \textbf{Homophily} & \textbf{$\delta$} \\
         \midrule
         TADPOLE & $1\,277$ & $0.7392$ & $1.31 \cdot 10^{-4}$ \\
         COVID & $65$ & $0.7569$ & $2.78\cdot 10^{-3}$ \\
         ABIDE ($k$=5) & $871$ & $0.6009$ & $1.92\cdot 10^{-4}$ \\
         Synthetic & $1\,000$ & varying &  $1.79\cdot 10^{-4}$ \\
         \bottomrule
    \end{tabular}
    \label{tab:homophily_delta}
\end{table}

\subsection{DP Training of GNNs on Population Graphs}
We summarise the results of non-DP and DP training at different privacy budgets in Table \ref{tab:results}.
\begin{table*}[h]
\centering
\scriptsize
\caption{Test set accuracy (\%) of non-DP and DP models at different $\varepsilon$-values across five random seeds.}
\begin{tabular}{lllllll}
    \toprule
    \textbf{Dataset} & \textbf{Non-DP} & \textbf{Sub-graphing}  & \textbf{DP ($\varepsilon=20$)}  &\textbf{DP ($\varepsilon=15$)} & \textbf{DP ($\varepsilon=10$)} & \textbf{DP ($\varepsilon=5$)} \\
    \midrule
    TADPOLE & 72.73 $\pm$ 1.39 & \textbf{76.09 $\pm$ 1.73} & 72.42 $\pm$ 0.94 & 71.02 $\pm$ 1.22  & 70.39 $\pm$ 0.43 & 69.45 $\pm$ 1.82\\ 
    Covid &  72.31 $\pm$ 11.51 & \textbf{73.85 $\pm$ 3.77} & 69.23 $\pm$ 8.43 & 69.23 $\pm$ 12.87 & 66.15 $\pm$ 10.43 & 56.92 $\pm$ 12.87 \\
    ABIDE ($k$=5) & 58.86 ± 0.81 & \textbf{65.14 ± 2.37}	& 57.83 ± 2.02 & 55.54 ± 2.62 & 53.71 ± 2.73 & 54.17 ± 2.97 \\
    ABIDE ($k$=30) & \textbf{68.51 ± 2.75}  & 65.83 ± 3.57 & 53.49 ± 4.27 & 53.37 ± 1.47 & 51.89 ± 4.40	 & 51.43 ± 3.47 \\
    \bottomrule
\end{tabular}
\label{tab:results}
\end{table*}

As expected, a higher privacy guarantee results in lower model performance.
For the TADPOLE dataset, a DP guarantee of $\varepsilon=20$ achieves performance comparable to non-DP training and even at $\varepsilon=10$, performance is only about two percent lower than non-DP results. 
We attribute this to the informative underlying graph structure of the TADPOLE dataset, which stabilises graph learning.
For the ABIDE dataset, non-DP performance is better for the graph structure that uses $30$ neighbours ($k$=$30$) compared to only $5$ neighbours.
However, larger neighbourhoods lead to more noise being added during DP-SGD training, which impacts the privacy-utility trade-off on this graph. 
%The Covid dataset suffers more severely from DP training and the results on the 
%ABIDE dataset with $30$ neighbours are close to random predictions at a privacy guarantee of $\varepsilon=20$. 
For all datasets apart from the ABIDE dataset with $30$ neighbours ($k$=$30$), the model trained without DP, but employing sub-graph sampling (\say{sub-graphing}) and gradient clipping out-performs the non-DP model trained without these techniques. 
We attribute this to the regularising effect of both aforementioned methods.
As seen, the ABIDE dataset has overall much lower accuracy and simultaneously, the homophily of the ABIDE dataset is the lowest among all datasets.
We therefore further investigate the impact of the homophily of the graph structure on the performance of DP population graphs in Section \ref{sec:impact_graph_structure}. 

\subsection{Membership Inference Attacks}
The dependencies between graph elements render GNNs more vulnerable to MIA \cite{liu2016dependence}.
Moreover, in the transductive setting of graph learning, test node features are included in the forward pass, which facilitates MIA \cite{he2021node}.
To empirically audit the privacy leakage of sensitive patient data from our GNN models, we employ the MIA implementation of Carlini et al. \cite{Carlini2021MembershipIA}.
We perform these experiments on the GNNs trained on the TADPOLE dataset, as it is known that higher model accuracy improves MIA success \cite{Carlini2021MembershipIA}.
The adversary/auditor in this membership inference scenario has full access to the trained model $f_\theta$, its architecture, and the graph, including its ground-truth labels \cite{Carlini2021MembershipIA}.
We trained $128$ shadow models to estimate the models' output logit distributions and create a classifier that predicts whether a specific example was used as training data for the model $f_\theta$. 

\begin{figure}
    \centering
    \begin{subfigure}[b]{0.48\textwidth}
    \centering
    \includegraphics[width=0.8\textwidth]{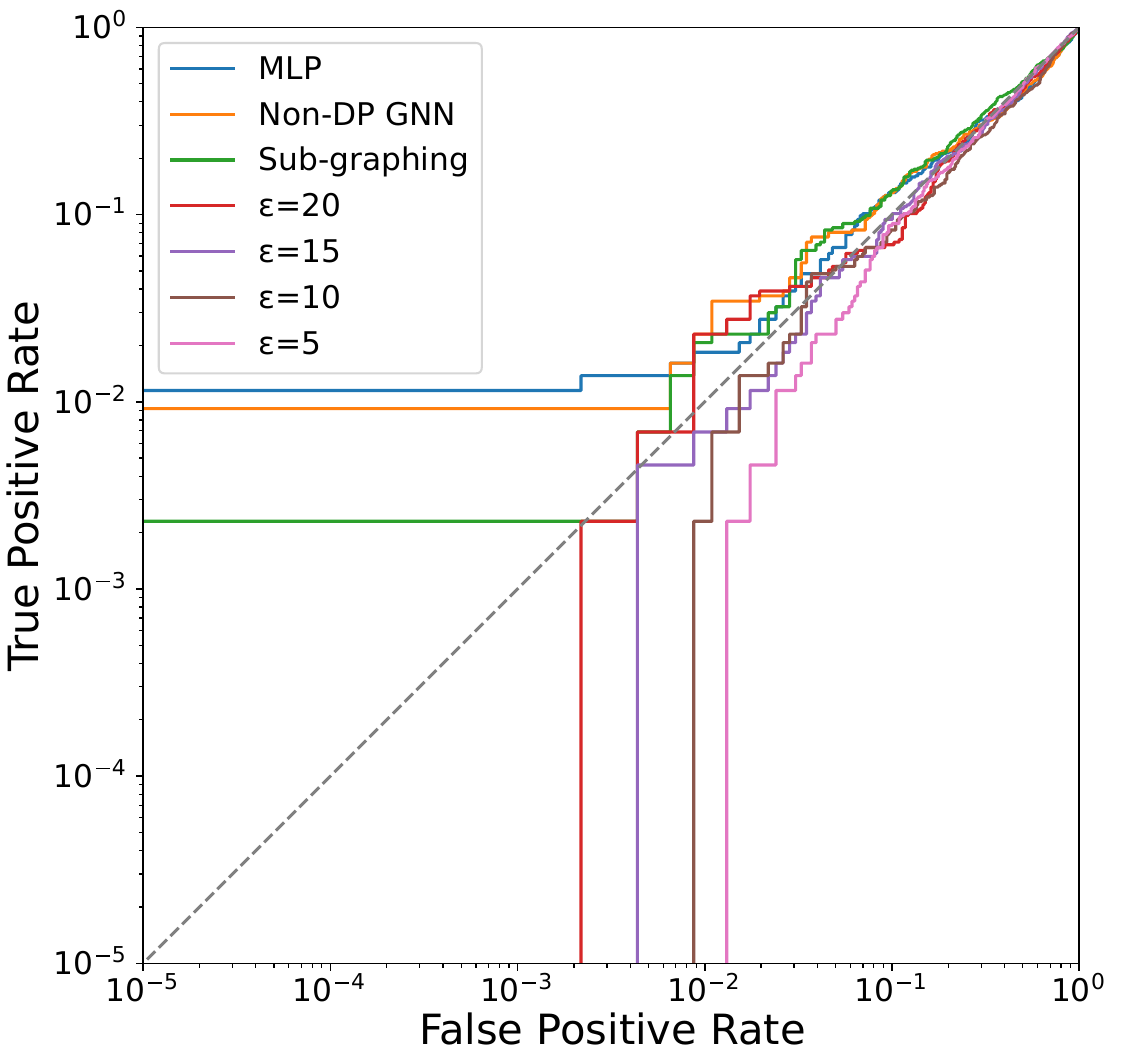}
    \caption{Empirical ROC curves for MIA on the TADPOLE dataset.}
    \label{fig:mia}
  \end{subfigure}
  \hfill
  \begin{subfigure}[b]{0.48\textwidth}
    \centering
    \includegraphics[width=\textwidth]{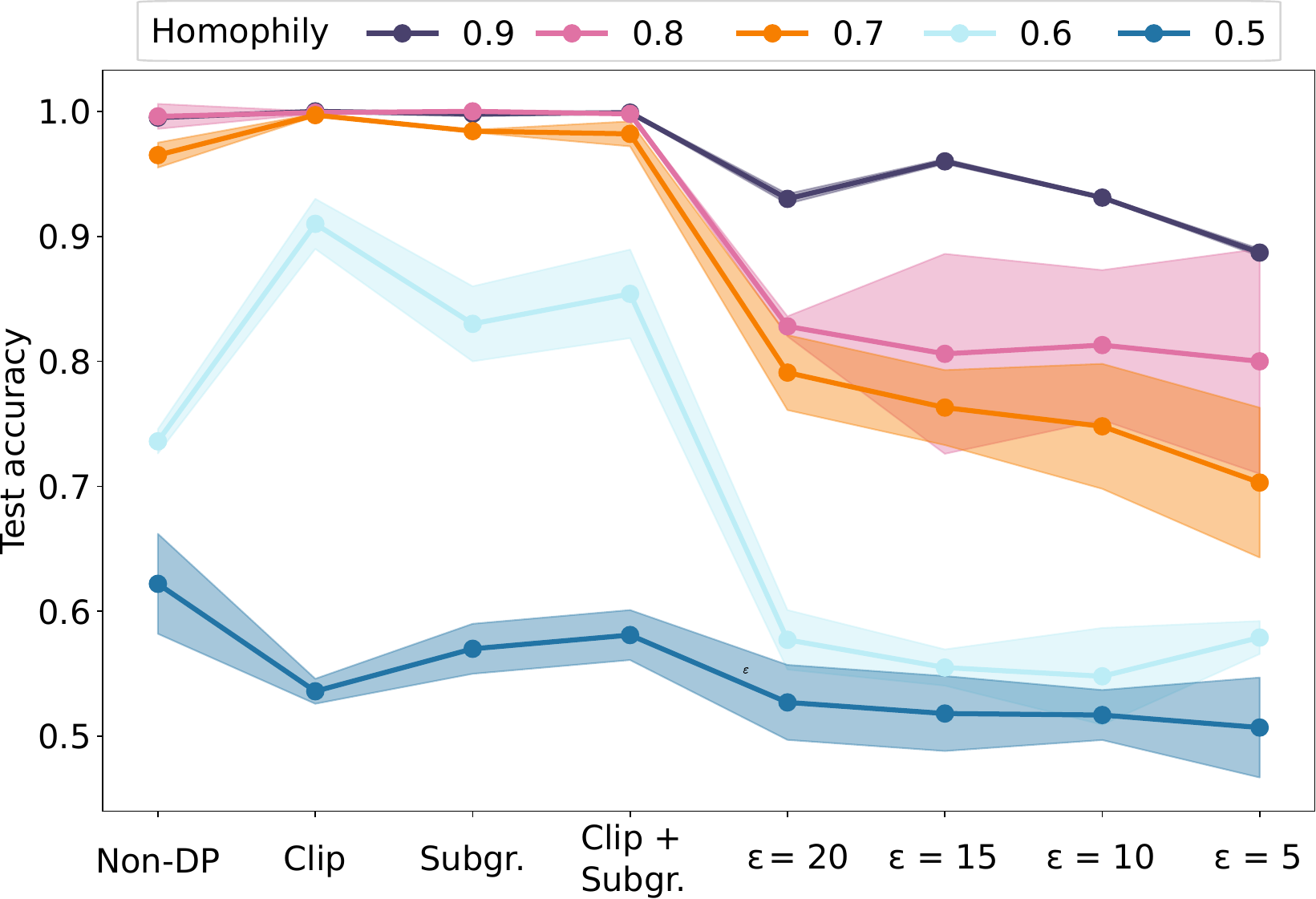}
    \caption{Impact of sub-graphing/ noise addition at different $\varepsilon$. \textit{Clip}: clipping only, \textit{Subgr.}: sub-graphing only.}
    \label{fig:synth}
  \end{subfigure}
  \caption{}
\end{figure}

In Figure \ref{fig:mia}, we report the log-scale receiver operating characteristic (ROC) curve of the attacks and report the true positive rate (TPR) at three fixed, low false positive rates (FPR) ($0.1\%, 0.5\%, 1\%$).
The attack's success rates are also summarised in Table \ref{tab:mia}.

\begin{table*}[h]
    \centering
    \scriptsize
    \caption{MIA results on the MLP and the non-DP and DP GNNs at different privacy budgets.}
    \begin{tabular}{lllclclc}
    \toprule
         \textbf{Model} & \textbf{Variant} & \multicolumn{2}{c}{\textbf{$\leq$ 0.001 FPR}} & \multicolumn{2}{c}{\textbf{$\leq$ 0.005 FPR}} & \multicolumn{2}{c}{\textbf{$\leq$ 0.01 FPR}}  \\
         & & \textbf{TPR} & $\mathcal{P}$ & \textbf{TPR} & $\mathcal{P}$ & \textbf{TPR} & $\mathcal{P}$ \\
         \midrule
         MLP & - & 0.0115 & - & 0.0138 & - & 0.0184 & - \\
         \midrule
         GNN & Non-DP & 0.0092 & - & 0.0092 & - & 0.0230 & - \\
             & Sub-graphing & 0.0023 & - & 0.0069 & - & 0.0207 & - \\
             & DP ($\varepsilon=20$) & 0.0000 & 1.0 & 0.0069 & 1.0 & 0.0230 & 1.0 \\
             & DP ($\varepsilon=15$) & 0.0000  & 1.0 & 0.0046 & 1.0 & 0.0069 & 1.0 \\
             & DP ($\varepsilon=10$) & 0.0000 & 1.0 & 0.0000 & 1.0 & 0.0023 & 1.0 \\
             & DP ($\varepsilon=5$) & 0.0000 & 0.1485 & 0.0000 & 0.7422 & 0.0000 & 1.0 \\
         \bottomrule
    \end{tabular}
    \label{tab:mia}
\end{table*}

Furthermore, we derive the maximum TPR (i.e. power) that is theoretically achievable for a given $(\varepsilon, \delta)$ setting through the duality between $(\varepsilon, \delta)$-DP and hypothesis testing DP \cite{dong2021gaussian}.
We will refer to this maximum achievable TPR as the adversary's \textit{supremum power} $\mathcal{P}$.

As seen, for FPR-values $<0.001$, the MIA is unsuccessful.
As the FPR tolerance is increased, models trained with weaker privacy guarantees ($\varepsilon \in \{20,15\}$) yield positive TPR when attacked, with TPR values approaching these of models trained without DP guarantees (\textit{Non-DP} and \textit{Sub-graphing} variants) in case of $\varepsilon=20$.
Interestingly, the model trained at $\varepsilon=5$ successfully resists membership inference even at an FPR value of $0.01$. 
Moreover, we observe that the GNN trained with clipped gradients is less vulnerable to membership inference than the GNNs trained without gradient clipping.
This is in line with the findings in \cite{Carlini2021MembershipIA} that clipping the gradients during training offers some (empirical) protection against MIAs.

\subsection{Impact of Graph Structure on Performance}\label{sec:impact_graph_structure}

The interaction between memorisation and generalisation in neural networks is of particular interest to the privacy community.
Feldman \cite{feldman2020does} hypothesises that especially noisy and atypical data from the long tail of the data distribution requires memorisation.
This increases the negative impact of DP training for those samples.
To investigate the applicability of the long-tail hypothesis to graph databases, we evaluate the impact of graph structure, measured in terms of homophily, on accuracy.
Concretely, we hypothesise that graphs with low homophily are \say{noisier} and therefore suffer more from DP training.
For example, at a homophily of $0.5$ on a binary node classification dataset, on average, half of all neighbours have the same label and the other half the opposite label.
Thus, when applying message-passing on such a graph, the node features will get averaged over an approximately equal number of nodes from both labels.
This makes it nearly impossible to learn meaningful node feature embeddings, such that learning likely relies nearly exclusively on memorisation.

The results using a synthetic dataset with different levels of homophily are summarised in Table \ref{tab:hom_exp} and visualised in Figure \ref{fig:synth}.

\begin{table}[ht]
\centering
\scriptsize
\caption{Results of the experiments on the synthetic dataset at different homophily values; All results refer to test accuracy (\%), with the best ones highlighted in bold. Compare Section \ref{sec:experiments} for details. Hom.: Homophily, Subg.: Sub-graphing.}
\begin{tabular}{lllllllll}
    \toprule
    \textbf{Hom.} & \textbf{Non-DP} & \textbf{Clipping} & \textbf{Sub-graphing} & \textbf{Subg. + Clip.}  & \textbf{DP ($\varepsilon=20$)}  &\textbf{DP ($\varepsilon=15$)} & \textbf{DP ($\varepsilon=10$)} & \textbf{DP ($\varepsilon=5$)} \\
    \toprule
    0.9 & 99.90 $\pm$ 2.00 & \textbf{100.0 $\pm$ 0.00} & 99.80 $\pm$ 0.40 & 99.90 $\pm$ 0.00 &  93.00 $\pm$ 4.17  & 96.00 $\pm$ 2.49 & 93.10 $\pm$ 1.36 & 88.70 $\pm$ 3.06 \\
        % 0.9 & 0.995 $\pm$ 0.002 & 0.9990 $\pm$ 0.002 &  0.9300 $\pm$ 0.0417 & 0.9600 $\pm$ 0.0249 & 0.93100 $\pm$ 0.01356466 & 0.88700 $\pm$ 0.03059 \\
    0.8 & 99.62 $\pm$ 0.37 & 99.90 $\pm$ 0.00 & \textbf{100.0 $\pm$ 0.00}  & 99.80 $\pm$ 2.45 & 82.80 $\pm$ 7.83 & 80.60 $\pm$ 10.8 & 81.30 $\pm$ 6.40 & 80.00 $\pm$ 8.76 \\ % 87.40 $\pm$ 1.428285
      %  0.8 & 0.9960 $\pm$ 0.006742 & 0.999 $\pm$ 0.002000 & 1.000 $\pm$ 0.00 & 0.9980 $\pm$ 0.00245 & 0.8280 $\pm$ 0.0783 & 0.8060 $\pm$ 0.1084 & 0.8131 $\pm$ 0.064 & 0.8000 $\pm$ 0.087579 \\ % 87.40 $\pm$ 1.428285
    0.7 & 96.50 $\pm$ 1.23 & \textbf{99.70 $\pm$ 0.40} & 98.42 $\pm$ 0.51 & 98.20 $\pm$ 0.50 & 79.10 $\pm$ 3.00 & 76.30 $\pm$ 3.10 & 74.80 $\pm$ 4.55 & 70.30 $\pm$ 5.64 \\
      %  0.7 & 0.96499 $\pm$ 0.01225 & 0.9969999 $\pm$ 0.004000 & 0.9841666 & 0.9820 $\pm$ 0.005 & 0.791 $\pm$ 0.030 & 0.763 $\pm$ 0.031 & 0.7480 $\pm$ 0.045453 & 0.703 $\pm$ 0.056444\\
    0.6 & 73.60 $\pm$ 0.97 & \textbf{91.00 $\pm$ 2.24} & 83.10 $\pm$ 3.10 & 85.40 $\pm$ 3.54 & 57.72 $\pm$ 2.38  & 55.50 $\pm$ 1.45 & 54.80 $\pm$ 3.87 & 57.90 $\pm$ 1.32  \\
          % 0.6 & 0.7360 $\pm$ 0.009695 &  0.9099999999999999, 0.022360679 & 0.8310 $\pm$ 0.031048 & 0.8540 $\pm$ 0.03541186 & 0.5772 $\pm$ 0.023791 & 0.5549999999 $\pm$ 0.01449 & 0.54799999 $\pm$ 0.038678 & 0.5790 $\pm$ 0.0131909  \\
    0.5 & \textbf{66.10 $\pm$ 1.361} & 53.60 $\pm$ 1.16 & 57.00 $\pm$ 1.76 & 58.10 $\pm$ 2.03 & 52.71 $\pm$ 3.34 & 51.82 $\pm$ 3.43 & 51.70 $\pm$ 2.38 & 50.70 $\pm$ 3.87 \\
    %     0.5 & 0.6610 $\pm$ 0.0135646599 & & 0.570 $\pm$ 0.01761 & 0.581 $\pm$ 0.0203 & 0.527 $\pm$ 0.0334 & 0.518 $\pm$ 0.0343 & 0.517 $\pm$ 0.02379 & 0.507 $\pm$ 0.038678 \\
    % 0.3 & 0.9620 $\pm$ 0.01077 & 0.8180 $\pm$ 0.02 & \\
    % 0.1 & 0.9970 $\pm$ 0.0040 &  \\
    \bottomrule
\end{tabular}
\label{tab:hom_exp}
\end{table}

The generalisation gap is especially large on low-homophily graphs, indicating over-fitting in the non-DP setting (not shown in the Table).
Model accuracy in the non-DP setting profits more from the regularising effects of clipping and sub-graphing in graphs with lower homophily ($0.6$) compared to high homophily ($0.9$).
We note that, in a binary classification task, homophily values are symmetric about $0.5$.

As expected, at the lowest homophily of $0.5$, learning is severely compromised without DP, and the regularising effect of clipping and sub-graphing actually harms accuracy.
Moreover, learning is nearly impossible with DP, corroborating that, in this setting, the accuracy benefit of non-DP learning is mostly due to memorisation.
In addition, under DP, graphs with high homophily ($0.9$) suffer a lower performance decrease compared to graphs with low homophily, likely due to the favourable graph structure for the learning task, i.e. not requiring strong memorisation.

\section{Discussion, Conclusion, and Future Work}
In this work, we investigate the practicality and challenges of differential private (DP) graph neural networks (GNNs) applied to medical population graphs. 
The utilisation of population graphs in medicine has shown promising improvements in performance for disease prediction \cite{parisot2017spectral} or age estimation \cite{cosmo2020latent}.
However, it comes with an additional challenge, namely the requirement of an explicit graph construction step.
This can lead to poor graph structures, regarding the homogeneity of neighbourhoods, which can be measured by the homophily metric.
Population graphs contain sensitive medical data of several subjects, which requires protection when applying DL methods to these graphs.
Applying DP to GNNs requires special formulations of DP concepts like privacy amplification techniques and DP-SGD methods \cite{daigavane2021node}.
We here evaluate privacy-utility trade-offs of DP GNNs trained on medical population graphs and reveal interesting correlations between the graph structure and performance of DP GNNs. 
When the underlying graph structure of a dataset has low homophily (indicating diverse neighbourhoods with different labels), DP has a stronger negative impact on model performance compared to datasets with high homophily.
This finding and its possible connection to the long-tail hypothesis \cite{feldman2020does} is a promising direction for future work to potentially improve DP methods for GNNs through improving the underlying graph structure.
Moreover, homophily is not the only measure for the \say{quality} of a graph structure.
Further metrics should be evaluated, which may shed more light on the impact of graph structure on the performance of DP GNNs.

\bibliography{literature}
\bibliographystyle{ieeetr}

\clearpage
\appendix

% \section{Datasets and graph construction}
% \textbf{Synthetic Dataset}
% 50 features, 5 informative, 1000 nodes, graph construction using $k$-NN, where $k \in \{... \}$. We use all features for graph construction and as node features.

% \textbf{ABIDE dataset} For the ABIDE dataset, we used sex and site as features for graph construction and the imaging features as node features

% \textbf{TADPOLE dataset} For the TADPOLE dataset, we use clinical features for graph construction and image-derived features as node features

% \textbf{Covid dataset}

\end{document}